\documentclass[lettersize,journal]{IEEEtran}
\usepackage{url}
\usepackage{amsmath,graphicx,cite}
\usepackage{amsfonts}
\usepackage{authblk}
\usepackage{algpseudocode}
\usepackage[ruled,linesnumbered]{algorithm2e}
\usepackage{verbatim}
\usepackage{longtable}
\usepackage{rotating}
\usepackage{float}
\usepackage{soul}
\usepackage{multirow}
\usepackage{tabularx}
\usepackage{bigints}
\usepackage{tikz}
\usepackage{multirow} 
\usepackage{amsmath,amstext,amsfonts,amssymb}
\usepackage{amsbsy}
\usepackage{amsthm}
\usepackage{diagbox}
\usepackage{mathabx}
\usepackage{mathrsfs}
\usepackage{bbm}
\usepackage{dsfont}
\usepackage{mathtools}
\usepackage{color}
\usepackage{caption}
\usepackage{subcaption}
\usepackage{morefloats}
\usepackage{float}
\usepackage{acronym}
\usepackage{ulem} 

\usetikzlibrary{arrows,shapes,positioning,shadows,trees}
\usetikzlibrary{decorations.pathreplacing}

\newacro{GenAI}[GenAI]{generative AI}
\newacro{LLM}[LLM]{Large Language Model}
\newacro{LTM}[Large-GenAI-Model]{Large-GenAI-Model}
\newacro{FM}[FM]{Foundation Model}
\newacro{AI}[AI]{Artificial Intelligence}
\newacro{LM}[LM]{Language Model}
\newacro{PTLM}[PTLM]{Pre-Trained Language Model}
\newacro{NLP}[NLP]{Natural Language Processing}
\newacro{DL}[DL]{Deep Learning}
\newacro{NN}[NN]{Neural Network}
\newacro{CNN}[CNN]{Convolutional Neural Network}
\newacro{RNN}[RNN]{Recurrent Neural Network}
\newacro{GNN}[GNN]{Graph Neural Network}
\newacro{ML}[ML]{Machine Learning}
\newacro{CV}[CV]{Computer Vision}
\newacro{SSL}[SSL]{Self-Supervised Learning}
\newacro{TL}[TL]{Transfer Learning}
\newacro{FM}[FM]{Foundation Model}
\newacro{NNLM}[NNLM]{Neural Network Language Model}
\newacro{LSTM}[LSTM]{Long Short-Term Memory}
\newacro{GPT}[GPT]{generative pre-trained transformer}
\newacro{BERT}[BERT]{Bidirectional Encoder Representation from Transformer}
\newacro{NLU}[NLU]{Natural Language Understanding}
\newacro{NLG}[NLG]{Natural Language Generation}
\newacro{T5}[T5]{Text-to-Text Transfer Transformer}
\newacro{ITL}[ITL]{In-Context Learning}
\newacro{RLHF}[RLHF]{Reinforcement Learning with Human Feedback}
\newacro{MHA}[MHA]{Multi-Head Attention}
\newacro{CLM}[CLM]{Causal Language Modeling}
\newacro{MLM}[MLM]{Masked Language Modeling}
\newacro{PLM}[PLM]{Permuted Language Modeling}
\newacro{DAE}[DAE]{Denoising AutoEncoder}
\newacro{RF}[RF]{radio frequency}
\newacro{QoS}[QoS]{quality-of-service}
\newacro{VAE}[VAE]{variational autoencoder}
\newacro{GAN}[GAN]{generative adversarial network}
\newacro{SON}[SON]{self-organizing networks}
\newacro{AI}[AI]{artificial intelligence}
\newacro{CSI}[CSI]{channel state information}
\newacro{FDD}[FDD]{Frequency Division Duplexing}
\newacro{MIMO}[MIMO]{multiple-input multiple-output}
\newacro{AoA}[AoA]{angle-of-arrival}
\newacro{AoD}[AoD]{angle-of-departure} 
\newacro{KPI}[KPI]{key performance indicator} 
\newacro{JSCC}[JSCC]{Joint Source-Channel Coding} 
\newacro{AGI}[AGI]{artificial general intelligence} 
\newacro{LLaMA}[LLaMA]{large language model Meta AI} 
\newacro{NeRF}[NeRF]{neural radiance field}

\begin{document}
\title{Large Generative AI Models for Telecom: \\ The Next Big Thing?}

\author{Lina Bariah, Qiyang Zhao, Hang Zou, Yu Tian, Faouzi Bader, and Merouane Debbah
\vspace{-30pt}
}

\maketitle

\begin{abstract}
The evolution of generative artificial intelligence (GenAI) constitutes a turning point in reshaping the future of technology in different aspects. Wireless networks in particular, with the blooming of self-evolving networks, represent a rich field for exploiting GenAI and reaping several benefits that can fundamentally change the way how wireless networks are designed and operated nowadays. To be specific, large GenAI models are envisioned to open up a new era of autonomous wireless networks, in which multi-modal GenAI models trained over various Telecom data, can be fine-tuned to perform several downstream tasks, eliminating the need for building and training dedicated AI models for each specific task and paving the way for the realization of artificial general intelligence (AGI)-empowered wireless networks. In this article, we aim to unfold the opportunities that can be reaped from integrating large GenAI models into the Telecom domain. In particular, we first highlight the applications of large GenAI models in future wireless networks, defining potential use-cases and revealing insights on the associated theoretical and practical challenges. Furthermore, we unveil how 6G can open up new opportunities through connecting multiple on-device large GenAI models, and hence, paves the way to the collective intelligence paradigm. Finally, we put a forward-looking vision on how large GenAI models will be the key to realize self-evolving networks.
\end{abstract}


\section{Introduction}

While recent wireless networks have witnessed several revolutions in terms of technological trends, it is apparent that future wireless generations are converging towards solidifying the principle of self-built, self-evolving networks, particularly with the maturing of the \ac{SON} paradigm and the remarkable advancements in \ac{AI} technologies \cite{chafii2023}. The core principle of \ac{SON} is based on the concept of enabling wireless networks to have the capability to adjust, reconfigure, and optimize their functions and parameters according to particular network conditions and user demands. However, despite the fact that \ac{SON} is aimed for realizing automation in wireless networks, its performance is dependent on predefined network conditions and their corresponding configurations. The ultimate vision of self-driven networks is to realize a retained network performance, maintain sustainable resilience to network variations, and design versatile networks that are capable of handling new network conditions and scenarios. 

As a cornerstone, \ac{GenAI} can be a key player in realizing the vision of self-evolving networks. \ac{GenAI} refers to a class of \ac{AI} that is designed to generate new content, including text, images, videos, etc., based on patterns and information learned from large datasets. Accordingly, \ac{GenAI} holds immense potentials in several directions, e.g., data augmentation, text and image generation, question-answering, sentiment analysis, conversational agents, human-machine interactions, automation, and many more. Within this context, \acp{LLM}, a foundation model built on natural language understanding and a sub-field of \ac{GenAI}, have recently attracted a considerable attention from the research community as a revolutionary technology in the field of AI \cite{zhao2023survey}. Subsequently, multi-modal \acp{LLM} have expanded the capabilities of \acp{LLM} from natural language to 2D/3D vision, sound signal, X-ray, etc., rendering them a promising candidate for applications in different domains including telecommunications \cite{zhang2023metatransformer}.

Among several \acp{LLM}, Falcon \ac{LLM} \cite{Falcon}, \ac{GPT}-2/3/4, \ac{BERT}, \ac{LLaMA}, as well as visual language models, e.g., DALL-E3 based on Contrastive Language-Image Pre-Training (CLIP) and diffusion model, have strongly impacted how AI is employed for inference and decision-making purposes, and laid down a new base for novel applications that can exploit the potentials of \ac{GenAI} models. This is rooted to the generative and predictability capabilities of these \acp{LTM}, in which large models (mainly based on the transformer architecture) are trained over a vast amount of unlabeled multimodal data (primarily textual and/or visual data), and therefore, are enabled to understand and generate human-like languages. Through the self-attention mechanism of transformers and the large amount of training data, the developed large models are able to  capture the statistical patterns and relationships in the provided data, and hence, to predict and generate the required data. Similar approaches apply to visual generative models, where \acp{VAE}, \acp{GAN} and diffusion models can be leveraged to map textual data with images and vice versa. 

Being a promising candidate to revolutionize several technologies in various fields, we believe that \ac{GenAI} models can introduce a radical change in wireless network design and operation. In particular, we anticipate that \acp{LTM} will introduce a tangible enhancement in the performance of different schemes within the Telecom domain. This can be achieved through exploiting the generative capabilities of \acp{LTM} in addition to the multimodality nature of the data acquired in wireless networks, including \ac{RF} signals, and 2D and 3D visual representations of wireless environments, to attain improved contextual, situational, and temporal awareness, and therefore, enhanced wireless communication. Furthermore, it is noted that \acp{LTM} can enable wireless networks to enjoy predictability features, and hence, realize improved and proactive localization, beamforming, power allocation, handover, as well as, spectrum management, even for unseen network scenarios. On the other hand, wireless network can connect multiple \ac{GenAI} models with novel communication mechanisms at semantic and effectiveness level, to provide faster sensing, inference, action with reduced resources consumption.

Recently, \ac{AI} has been playing a key role in a wide range of applications in Telecom networks. 
However, most \ac{AI} solutions in wireless networks are developed for solving dedicated problems, which is inefficient for general use in wireless network with emerging applications. \Acp{LTM} on the other hand represent a pre-trained model that can solve many downstream and upstream tasks with efficiently designed prompts or through fine-tuning, rendering such models a promising candidate for \ac{AI} native networks, in which massive data pertinent to system design (architecture, protocol) and operation (signaling, procedure) can be used with \acp{LTM} to perform many tasks cross different layers.
We foresee that \ac{LTM}-like solutions could generalize over unseen data, tasks, and scenarios, and reduce the complexity, cost, and power in network design, deployment and operation. Large GenAI models in Telecom are envisioned to have a pivotal role in enhancing wireless communications by enabling automatic network control and optimization, as well as facilitating complex tasks like channel estimation and environment reconstruction. Additionally, the integration of cloud-based GenAI models into network edge devices should prompt a shift in wireless communication towards a knowledge-based paradigm, promoting collaborative interactions among on-device GenAI models to improve resource efficiency and reduce energy consumption.
\begin{figure}
	\centering
	\resizebox{0.8\linewidth}{!}{\includegraphics{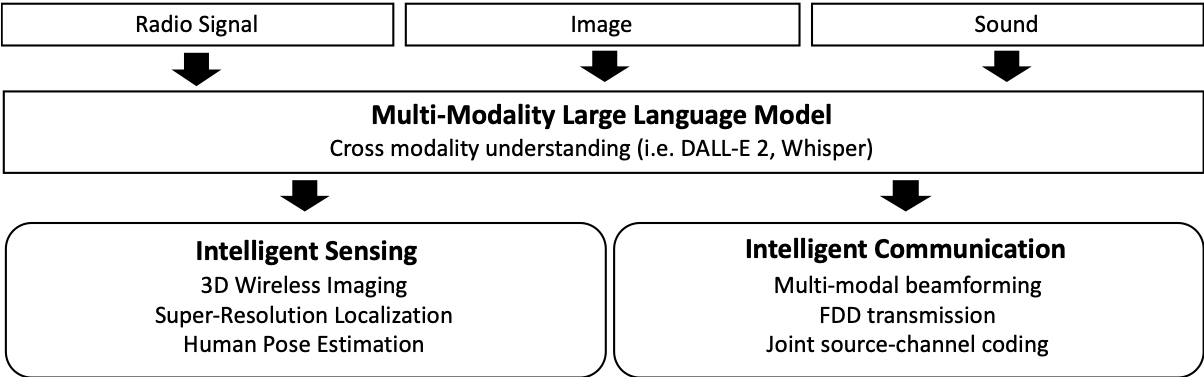}}
	  \caption{Example applications of \acp{LTM} for wireless sensing and communication}
    \label{fig:Sensing}
\end{figure}

\subsection{Contribution}

While there are several articles that discussed foundation models in general, and \acp{LLM} in particular, e.g., \cite{min2021recent,zhou2023comprehensive,hadi2023survey, koh2023grounding} - and the references therein, all of these articles have focused on the application of \acp{LLM} in the general domain, with specific interest in textual data with minor discussions on visual data. In this article, \textit{for the first time in the literature}, we explore the integral role that large generative models will play in future wireless networks. From the one hand, we aim to unveil the potentials of leveraging large generative models to pre-train a single foundation model that can subsequently be employed across a range of different downstream tasks in the Telecom domain, for improved wireless sensing and transmission. This approach has several significant advantages, including improved efficiency, reduced training requirements, and enhanced adaptability to varying network conditions, as well as the generalizabilty of these models in different propagation environments. For this purpose, we introduce a new term, namely, \textit{\Acp{LTM}}, which refers to large generative models that are designed and trained to fit various applications and use-cases of Telecom networks, from network design and configuration, to optimization and operation. On the other hand, we draw a roadmap to how wireless networks can potentially contribute to developing efficient large models, that are capable of accommodating the needs of future 6G networks. We finally reveal how the two paradigms (\ac{LTM} for wireless and wireless for \ac{LTM}) pave the way to the conceptualization of the \ac{AGI}-empowered wireless networks, the seed to fully self-evolving networks.


\section{\Ac{LTM} for Wireless}

\subsection{Large Language Models for Sensing}
\label{Sec:Localize}
\subsubsection{3D Wireless Imaging}
 
\Ac{DL} models have fundamentally contributed to the development of improved wireless sensing schemes, in which \ac{RF} data can be acquired and mapped to 2D images for sensing applications, including localization, remote sensing, and resource allocation. While \ac{DL} approaches have demonstrated an acceptable performance in several scenarios, they lack the generalizability to new network requirements and scenarios. Furthermore, they suffer from the complexity that arises from the reliance on supervised training data. For the latter, to achieve high-resolution sensing for mission-critical applications, large labeled data-sets are required, rendering the labeling process a huge burden. Within the same concept, \textcolor{black}{\ac{NeRF} framework \cite{mildenhall2021nerf} was proposed to understand the interaction of light signals with objects in a scene to create realistic and detailed 3D representations, relying on the \ac{DL} architecture. Also, NeRF2 was proposed to understand the physical wave transmission which can be leveraged for accurate CSI acquisition and hence, improved performance in localization applications, however, it suffers from high computational complexity and lacks the ability to scale.} The recent evolution in visual \ac{GenAI} models has opened the horizon to a new era of sensing capabilities, in which machines now have the potential to generate high-quality images and understand visual content, i.e., generating 2D and 3D images from textual description or mapping images with corresponding text. 

Among others, DALL-E3 (capable of generating high-quality images from textual descriptions), CLIP (capable of matching images with text captions) , Vector Quantized GAN (generate diverse and high-quality images), and GPT-4V (focuses on generating textual descriptions of images), have been identified as core models for 2D and 3D image generation from text, or vice versa. These models rely on the several advanced techniques including conditional \acp{GAN}, discrete \acp{VAE}, and diffusion models to capture the visual patterns and their contextual relationship with textual content. 

The promising potentials of visual \ac{GenAI} models introduce a plethora of benefits that can be exploited for wireless communications, for enhanced system design and optimization. This vision is built upon the principle of enabling \ac{GenAI} models to understanding the cross features between different data modalities, including images of wireless propagation environments and \ac{RF} signals. Specifically, these visual models will be developed to generate super-resolution 3D images of the surrounding environment from measured wireless data. The constructed 3D image will be then leveraged to enable improved communication schemes, in which enhanced contextual and situational awareness can be realized, and therefore, achieving improved beamforming, handover, resource allocation, etc. This is due to the fact that Telecom visual \ac{GenAI} models are anticipated to capture the relationships between \ac{RF} data and different information that can be extracted from 3D images (static system topology, non-moving objects, dynamic objects information), and the impact of the latter on the former, enabling accurate multi-dimensional reconstruction of the wireless environment from the acquired RF signals. Note that it has been shown that visual GenAI model (i.e. diffusion model) can effectively perform image construction from brain signals (via functional magnetic resonance imaging (fMRI)) \cite{takagi2023high}. Moreover, meta-transformer has demonstrated a unified transformer framework to incorporate sound, X-ray, and hyper-spectrum signals to perform many downstream tasks such as segmentation, detection, provided that it can align different modalities on a semantic latent space \cite{zhang2023metatransformer}. Unlike conventional visual \ac{GenAI} models that map images with texts using text/image encoders/decoders, the architecture of Telecom visual \ac{GenAI} models are aimed to comprise encoders that are capable of extracting features and patterns from wireless signals and map these features into 3D images. Telecom-oriented \ac{LTM} framework can be developed to extract and concatenate images and signals embeddings, and allow the machine to understand the relationship between the two data modalities through self-attention mechanism.

\subsubsection{Super-Resolution Localization}

Super-resolution localization is an essential element in current wireless networks, since determining the accurate locations of different wireless devices or nodes in the network readily facilitates optimizing network performance, efficiently allocating resources, and ensuring maximized \ac{QoS} by dynamically adjusting network parameters, to accommodate the needs of different users according to their current status and positions. Current localization techniques follow mainly two directions, \ac{RF}-based localization, which rely on RF signals from anchor nodes for position detection, or vision-aided localization, which exploits visual data and image processing for feature extraction and object detection. While the latter can provide more contextual information that render higher localization accuracy, the limited field-of-view and calibration and precise alignment of cameras or sensors constitute limiting factors in such approaches. These challenges are further pronounced in multimodal localization, in which the fusion and alignment of different data modalities results represent a bottleneck in exploiting the full potential of multimodal localization.

Within this context, \ac{LTM} can be a game changer in enabling efficient multimodal localization schemes, in which the general and self-attention nature of these large models can be the key to detect the contextual and situational information of network users and nodes, capture the mutual displacement between multiple images, and cross-correlate these images and their variations with the corresponding electromagnetic behaviour of wireless signals. The benefits of utilizing \acp{LTM} for high-precision localization are two-fold. First, \acp{LTM} can introduce improved perception skills of the surrounding environment through integrating multimodal data that allows the understanding of the environmental, temporal, and situational aspects of the network events and behaviours, and their impact on the network performance. With the generative capabilities of \acp{LTM}, it is envisioned that a pretrained large model will be capable of accurately locating multiple users within the network. Second, the exploitation of \acp{LTM} in localization applications is stemmed from the fact that pretrained large models allow wireless networks to enjoy improved predictability skills. In more details, the integration of 3D images with \ac{RF} data enables the large models to identify both idle and active users, as well as predict their future activities through their generative and prediction capabilities. This offers a comprehensive understanding of the wireless environment and its variations, and accordingly enable enhanced network configuration and optimization. This yields more proactive networks, paving the way to self-built and self-healing networks. 

\subsection{Large Language Models for Transmission}

\subsubsection{Multi-modal beamforming}
The evolution of high-frequency communications, e.g., mmWave and THz communication, has given a rise to several issues pertinent to the transceiver design (including efficient high-power antenna design and beamforming/MIMO schemes), as well as the reliability of signals transmitted over high frequencies. This is due to the lossy and highly directional nature of high-frequency communication, which yields severe signals degradation even with minor blockage or beam misalignment. Accordingly, efficient beamforming and beam alignment mechanisms are essential to be developed to achieve reliable high-frequency communication. Typically, beam selection is performed through the reliance on a predefined beam codebook, in which a wide-range of beam sweeping is performed between the transmitter and the receiver in order to select the optimum beam to establish a reliable link between the two nodes. Such a procedure is usually associated with high beam training overhead, introducing a performance limiting factor in high-mobility and latency-sensitive applications. 

\Acp{LTM}, pretrained over a large data-set of beamforming scenarios, have the capability to predict the optimum beam that maximizes the signal strength and minimizes interference. This can be achieved through exploiting multimodality for providing additional information in regard to blockage probability and users status and activities, and hence, an \ac{LTM} can be used to capture the various features and patterns within the network dynamicity, and predict the optimum beam according to current and future network scenarios. 

\subsubsection{\ac{FDD} Transmission}
\begin{figure}
	\centering
	\resizebox{0.7\linewidth}{!}{\includegraphics{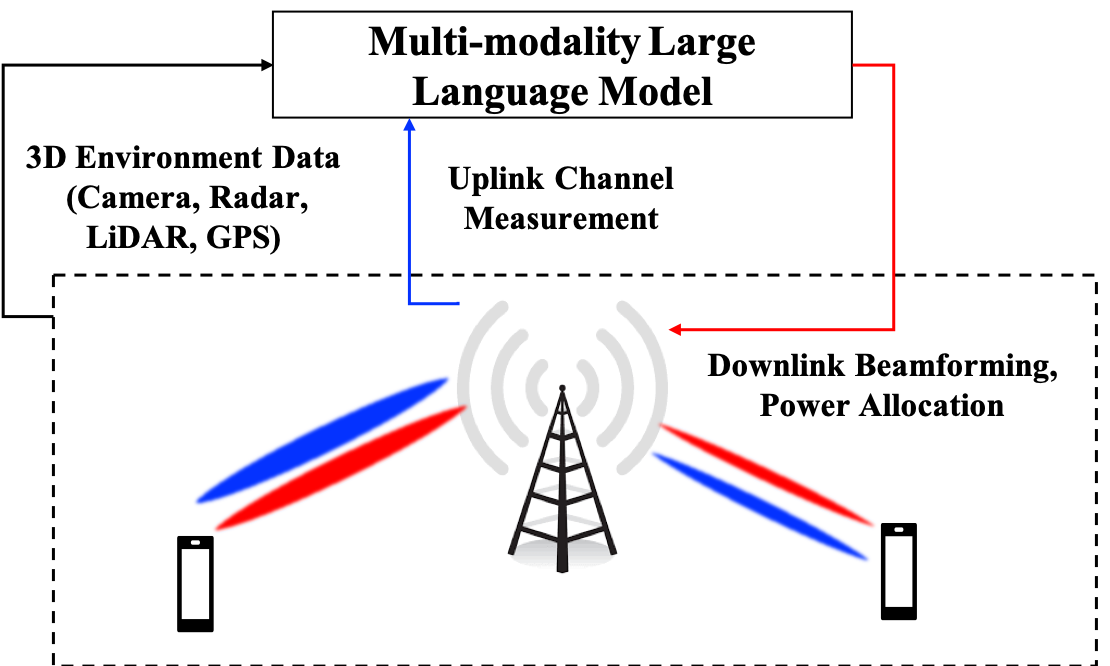}}
	  \caption{\Acp{LTM} for Beamforming in \ac{FDD} Systems}
    \label{fig:Beamforming}
\end{figure}

Within the context of \ac{FDD}, \acp{LTM} can be utilized for \ac{CSI} estimation purposes between the uplink and downlink transmissions. In conventional \ac{FDD} systems, where separate frequency bands are allocated for uplink and downlink transmissions, \ac{CSI} acquisition is typically performed separately for the two directions, consuming the network resources and introducing high latency. This issue is more pronounced in massive \ac{MIMO} scenarios, where it is very challenging to acquire all \ac{CSI} over the uplink \cite{yang2020deep}. Alternatively, partial uplink CSI knowledge can be exploited to extrapolate the full downlink \ac{CSI}. Through the self-attention mechanism and the generative capabilities of \acp{LTM}, we envisage that \acp{LTM} will be able to capture the inherent relationship between the uplink and downlink transmission and exploit 3D multimodal environment data (including camera, radar, LiDAR, and GPS) in order to select the optimum uplink and downlink beam pair that yields perfect alignment between the angle-of-arrival and angle-of-departure, at a particular user position (Fig. \ref{fig:Beamforming}). It should be emphasized that super-resolution localization achieved by \acp{LTM} (as discussed in Sec. \ref{Sec:Localize}) can directly impact the beamforming performance in \ac{FDD} systems, in which accurate localization allow \ac{LTM}-enabled beamforming schemes to predict future user position, and therefore, enable improved planning for the spectrum resource over the uplink and downlink transmissions. 

\subsubsection{\ac{JSCC}}
Conventional wireless networks are designed in a block-based architecture, where each process has a dedicated block, including channel estimation, coding, decoding, equalization, to name a few. With the identified requirements of future 6G networks, such architecture cannot achieve the envisioned key performance indicators of latency, reliability, spectral and energy efficiency, and connectivity. It is further emphasized that block-based architectures are difficult to scale to large, more generalized scenarios \cite{wang2022transformer}. In this regards, current research initiatives have been conducted to explore the appealing benefits of integrating channel and source coding into semantic-aware \ac{JSCC}, in order to serve use-cases with extreme latency and bandwidth requirements, that are computationally demanding in long block-length techniques. When incorporating different data modality, e.g., images and videos, to be transmitted, \ac{DL} approaches have demonstrated outstanding performance in terms of latency and quality-of-compression. However, \ac{DL} models rely heavily on a huge amount of labeled training data for the joint optimization of source and channel coding, which, in most scenarios, is not available. This readily impact the generalizability of the trained models to unseen or varying (channel conditions, interference, noise, etc.) network scenarios. 

 In this regard, \acp{LTM}, can facilitate the realization of efficient \ac{JSCC} schemes for improved wireless communication. In addition to the labeled data independency, \acp{LTM} can be exploited to understand the statistical behavior of the source data and to allow the reliable extraction of the needed information, and hence, realize efficient data compression \cite{gilbert2023semantic}. Furthermore, through the self-attention mechanism, \ac{LTM} introduce improved robustness to channel errors, through enabling efficient error correction mechanisms. This can be achieved by using \acp{LTM} to capture the inherent features of wireless channels behavior, error statistics, and source data, and then learn the relation between the source data and the corresponding channel coding requirements, and therefore, realize improved prediction for errors and promotes the development of robust error correction codes. Moreover, by understanding the long interdependencies between the source data and channel coding, adaptive mechanisms can be designed with reliance on \acp{LTM}, where real-time adjustment of the code rate, modulation, and code selection can be performed according to the current channel condition, striking a balance between source coding and channel coding performance.   


\section{Wireless for \Ac{LTM}}

\subsection{6G with Collective Intelligence}

An evolution of \ac{GenAI} is to empower a massive number of wireless devices to deliver collective intelligence. To achieve this, communication systems should be transferred from data and model-based to knowledge and reasoning-based paradigms. Conventional communication networks aim to transfer data from one network node to another under targeted \acp{KPI}. This is inefficient for a network connecting large-scale devices powered by \acp{LTM}. The goal of 6G is to develop a computing fabric that moves knowledge within the network. This can push nowadays cloud \ac{LTM} trained on world knowledge towards distributed collective intelligence. To achieve this, \acp{LTM} should be grounded to the real-world context, communicate with knowledge, to perform multi-agent planning, decision making, and reasoning. In this section, we summarize three key research perspectives to realize a multi-agent \ac{GenAI} wireless network. 

\subsubsection{Semantic communication}

\acp{LTM} built on raw data could cause large computation overhead and communication redundancy. Learning an effective semantic abstraction of raw data is essential to deploy and connect large \ac{GenAI} models on resource constrained wireless devices. This requires wireless physical layer to be redesigned in a semantic and effective manner for the \ac{GenAI} models. \Acp{LTM} can learn from the semantic compression and abstraction of raw data represented as knowledge \cite{gilbert2023semantic}. In this context, information should be characterized with a minimal structure, which is robust against changes in distribution, domain, and context. This does not only reduce the size of data and model in the device memory, but also can represent different data modalities with a common latent (concept) space. Knowledge can be potentially modeled on a topological space, such as graphs, simplices, cells, complexes, which can largely reduce the data transmitted between agents while being effective for \acp{LTM} to accomplish the targeted tasks. On the other hand, the higher-order interactions between concepts on topological space allows \acp{LTM} to perform logical reasoning from one domain to another, which reduces the needed raw data collection and exchange between agents.

\subsubsection{Emergent protocol learning}

The \ac{GenAI} powered wireless devices will have many emerging applications in different domains. Conventional 5G system protocols developed for specific applications (i.e. XR, MTC, IoT) are inflexible to meet the diverse goals of \ac{GenAI} use cases. An emergent, adaptable, and autonomous communication protocol on a goal-oriented basis is essential to allow multiple wireless \ac{GenAI} agents interact or collaborate effectively.

Multi-agent reinforcement learning (MARL) has the potential to learn a semantic language between agents through collaboration, such as when, where, and what information should be delivered. This permits to design a goal oriented, compositional and grounded wireless MAC and network layer protocols between devices. Moreover, multi-agent game can be an alternative technique to improve agent's action in a competition scenario. This can be performed jointly at the network level to optimize resource allocation, and at the device level to optimize vehicle or robotic control. Furthermore, \acp{LTM} can guide MARL and games using its expert model for planning, for improved convergence and reduce data collection.



\subsubsection{Distributed \acp{LTM} powered AI agent}

To achieve full autonomy for a self-evolved network, an autonomous agent with the capability of grounding, planning, criticising, and reasoning is essential. With autonomous agents, wireless devices not only perform inference with \acp{LTM}, but also perceive environment, plan tasks, memorize experiences and evaluate actions. This could lead to a fully autonomous control in both network and device without human intervention. To achieve this, 
the \ac{LTM} knowledge should be grounded to the real-world context. Reinforcement learning has recently been applied on \acp{LTM}, leveraging environment perception and rewards to improve the decisions made by the \acp{LTM} \cite{carta2023grounding}. With distributed planning, \acp{LTM} can a create sequences of actions to accomplish a task, and distribute sub-tasks to different wireless devices so as to improve the execution efficiency via cooperation. 
As a generative model, an LLM can make wrong decisions and answers, where the criticism between multiple LLM instances is proved to improve the performance \cite{du2023improving}. Moreover, \acp{LTM} can perform reasoning via planning, such as chain or tree of thoughts. By learning the logical relations between concepts abstracted from data, \acp{LTM} can have better generalization capabilities, with less training data, computation cost, and faster inference, which is important on energy constrained wireless devices. Furthermore, semantic communication and emergent protocol can be used to connect multiple autonomous agents effectively for cooperative planning and reasoning. The cloud-based \acp{LTM} can act as a world model guide wireless agents to optimize in a real-world context from \ac{LTM}'s common-sense knowledge.

\subsection{Use Cases of Collective Intelligence}

GenAI has the potential to realize automation in intent-driven networking. With Telecom \ac{LTM}-empowered autonomous agents, network can self-evolve in a distributed manner without human intervention. With the capability of distributed planning, \acp{LTM} can break down a higher-level intents into lower level actionable tasks in wireless network, sense radio environment, take action to configure system, and memorize experience for future. Furthermore, effectiveness communication between multiple \ac{GenAI} agents in network devices can eliminate the need for a centralized control, and reduce the signaling load on control plane.

Wireless GenAI network can also bring collective intelligence to autonomous vehicles. With autonomous agents deployed on vehicles and road side units, semantic communication and emergent protocols on the V2X data plane can largely improve the transmission efficiency, latency and reliability in both communication and control, compared to 5G delivering raw perception via eMBB and control commands via URLLC. Moreover, cooperative perception and control through \ac{LTM}-guided MARL or games can improve the traffic flow and safety, bringing full autonomy to vehicle network.

\section{AGI-Empowered Wireless Networks: A Vision Forward}

The evolution of several \ac{AI} paradigms is approaching the concept of \ac{AGI}, in which machines will enjoy a level of intelligence that is equivalent to or surpassing humans intelligence. The notion of the \ac{AGI} represents a paradigm for building machines that are capable of mastering a wide range of tasks without being particularly trained on them, through possessing a broad spectrum of human-like cognition capabilities, including understanding, learning, applying knowledge, and performing reasoning and inferring in various domains, in an autonomous fashion. It is envisioned that \acp{LTM}, with their generalized and generative capabilities, will be the keystone towards the successful deployment of \ac{AGI} in wireless networks, in which \ac{AGI}, empowered by \acp{LTM}, can be the orchestrator to the efficient planning, design, deployment, configuration, and operation of future wireless networks. In what follows, we put our forward-looking vision of how \acp{LTM}, or more generally foundation models, will be integrated into wireless networks, with the aim to realize the true vision of network automation.

\subsection{Task-Agnostic Large Telecom Model}

\textcolor{black}{The main purpose of adapting the large generative AI models into the Telecom domain is to develop foundation models that are trained using multi-modal data, in order to perform general Telecom tasks. These foundation models are aimed to enable improved communication and control in wireless networks. In particular, these foundation models, or large Telecom models, are anticipated to act as a general-purpose backbone to the network, with high scalability and flexibility to be deployed on edge devices. This can be achieved by designing large GenAI models to perform several general Telecom-oriented tasks, in order to reduce the cost of training multiple \ac{AI} models to perform specific tasks. After that, the pretrained model can be tailored to fit within a particular downstream task of interest, including modulation, coding, power allocation, beamforming, etc., through fine-tuning with the relevant data, through fine-tuning the foundation model using the relevant data. This represents a stepping stone to implementing \acp{LTM} at edge devices, where the fine-tuning process is much less complex and smaller data-set is required.}


\subsection{Self-Evolving Networks}

\begin{figure*}
	\centering
	\resizebox{0.85\linewidth}{!}{\includegraphics{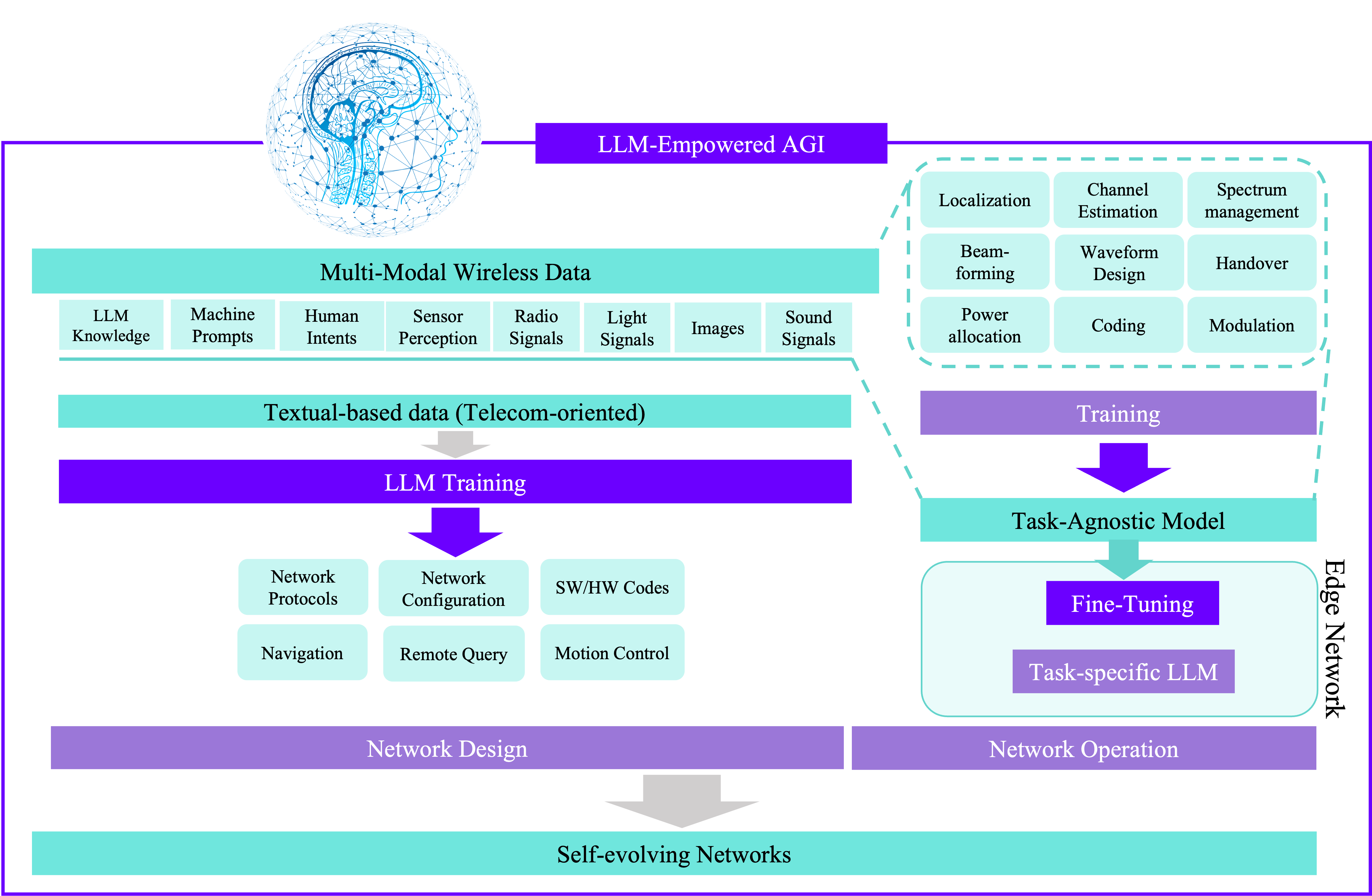}}
	  \caption{\ac{AGI}-empowered wireless networks.}
    \label{fig:Brain}
\end{figure*}

The concept of self-evolving networks refers to the networks that are capable to adapt, change, and evolve with the variations experiences in the network, and the surrounding environment. It can autonomously plan long-term tasks, make decisions, reason, and memorize logic, to achieve particular goals without human intervention. By leveraging \ac{AI}, these networks will have the capability to optimize their configurations in real-time fashion, according to the users demands and the network dynamics. This will result in a sustainable performance without human intervention. With the generalization capability, multi-modality, and planning and reasoning, \acp{LTM} can realize the vision of self-evolving networks which will go beyond the self-adaptation and self-optimization principles. In specific, we anticipate that \acp{LTM} will be leveraged to contribute to the initial steps of designing, planning, deploying, configuring, and operating wireless networks. The use of textual documents from the standards and research reports will allow pretrained \acp{LTM} to generate the required software codes and hardware design specification, which will be then go to the deployment stage. In the later stages, \acp{LTM} can be employed to identify the optimum configuration of the network according to the initial design, requirements, and users needs. It can be further exploited to build new communication schemes, that are not necessarily compatible with a particular standard, according to a particular network condition, for example, novel waveform to cope with high mobility scenario, modulation and coding design to accommodate multiple high data-rate users, etc. This vision of \textit{\ac{AGI}-empowered wireless networks} (Fig. \ref{fig:Brain}) can be achieved with the utilization of the diverse Telecom data modality and the predictability and generative nature of \acp{LTM}.

\section{Challenges and Open Research Directions}

\subsection{Telecom \acp{LTM} Architectures}

Despite the high number of architectures designed for the \ac{LTM}, that are capable of handling multimodality, adapting the paradigm of \acp{LTM} to the Telecom domain will require sophisticated architectures that are designed and trained from scratch, in order to fit within the Telecom domain. This is stemmed from the fact that Telecom data enjoy unique characteristics, compared to textual data and images/videos. In particular, it is expected that the available \ac{LLM} architectures are incapable of handling \ac{RF} data, and integrating it with other data modalities. 



\subsection{Distributed GPT}
With the vast amount of data generated at different nodes in the network, distributed GPT, in which the large generative AI model is trained in a distributed fashion, offers several benefits that make it appealing particularly for wireless networks. From the one hand, distributed GPT allows high horizontal scalability in which a larger number of users can contribute to the training process, and hence, reduced latency can be achieved. The latter can be a consequence of processing the data and training the model closer to the data source. Furthermore, distributed GPT can be more efficient in adapting to the dynamicity of wireless networks, in which decisions can be made faster according to locally-processed data. However, distributed GPT faces several challenges in terms of models coordination and aggregation, and data synchronization.

\subsection{Explainable Large GenAI Models in Telecom}
The explainability of large generative models is a key factor to be thoroughly studied in the Telecom domain. As being integrated into the network design, management, and control, and being granted authoritative and decision-making roles, it is essential to understand how these complex generative models arrive at their conclusions, and rationale behind their decisions. The explainability of \acp{LTM} is essential for two main reasons. First, in many scenarios, human operators work in collaboration with \ac{AI} agents, and therefore, it is important for these operators to understand why a particular decision is made by the \ac{AI} agent, enabling them to make more informed choices. Second, understanding the logic behind opens up many opportunities for optimization and enhancement, as well as realize trustworthy behavior for \ac{LTM} agents.

\subsection{GenAI On-Device}
The distributed nature and the emergence of edge-computing paradigm push for the deployment of large \ac{GenAI} models on edge devices, offering several benefits in terms of reduced latency, offline operation, as well as the reduce reliance on cloud services. However, taking into consideration the extremely large sizes of \acp{LTM} and the huge amount of data to be processed, computing and energy resources are two main factors to be optimized when implementing \ac{GenAI} models on edge devices. Several techniques can be adopted to optimize the model efficiency in order to reduce the computational requirements, including model compression and quantization, and knowledge distillation. Furthermore, such a resource optimization problem can be achieved through exploiting schemes such as tasks offloading, dynamic resource allocation, and optimized communication protocols.

\section{Conclusion}
In this article, we explored how \acp{LTM} can be an essential tool in designing, configuring, and operating future wireless networks. In particular, we identified the key opportunities, with respect to sensing and communication, that can be acquired when employing \acp{LTM} in wireless networks, and we overviewed the role of wireless networks in enabling machines to communicate using \acp{LTM}. Moreover, we laid down the foundation for the development of the \ac{AGI}-empowered wireless networks through \acp{LTM}, which paves the way to the successful implementation of self-evolving networks. 

\bibliographystyle{IEEEtran}
\bibliography{IEEEabrv,Refs}

\begin{thebibliography}{10}
\providecommand{\url}[1]{#1}
\csname url@samestyle\endcsname
\providecommand{\newblock}{\relax}
\providecommand{\bibinfo}[2]{#2}
\providecommand{\BIBentrySTDinterwordspacing}{\spaceskip=0pt\relax}
\providecommand{\BIBentryALTinterwordstretchfactor}{4}
\providecommand{\BIBentryALTinterwordspacing}{\spaceskip=\fontdimen2\font plus
\BIBentryALTinterwordstretchfactor\fontdimen3\font minus \fontdimen4\font\relax}
\providecommand{\BIBforeignlanguage}[2]{{%
\expandafter\ifx\csname l@#1\endcsname\relax
\typeout{** WARNING: IEEEtran.bst: No hyphenation pattern has been}%
\typeout{** loaded for the language `#1'. Using the pattern for}%
\typeout{** the default language instead.}%
\else
\language=\csname l@#1\endcsname
\fi
#2}}
\providecommand{\BIBdecl}{\relax}
\BIBdecl

\bibitem{chafii2023}
M.~Chafii \emph{et~al.}, ``Twelve scientific challenges for {6G}: Rethinking the foundations of communications theory,'' \emph{IEEE Commun. Surveys Tuts.}, 2023.

\bibitem{zhao2023survey}
W.~X. Zhao \emph{et~al.}, ``A survey of large language models,'' \emph{arXiv preprint arXiv:2303.18223}, Sep. 2023.

\bibitem{zhang2023metatransformer}
Y.~Zhang \emph{et~al.}, ``Meta-transformer: A unified framework for multimodal learning,'' Jul. 2023.

\bibitem{Falcon}
\BIBentryALTinterwordspacing
Falcon {LLM}. [Online]. Available: \url{https://huggingface.co/tiiuae}
\BIBentrySTDinterwordspacing

\bibitem{min2021recent}
B.~Min \emph{et~al.}, ``Recent advances in natural language processing via large pre-trained language models: A survey,'' \emph{ACM Computing Surveys}.

\bibitem{zhou2023comprehensive}
C.~Zhou \emph{et~al.}, ``A comprehensive survey on pretrained foundation models: A history from bert to chatgpt,'' \emph{arXiv preprint arXiv:2302.09419}, May 2023.

\bibitem{hadi2023survey}
M.~U. Hadi \emph{et~al.}, ``A survey on large language models: Applications, challenges, limitations, and practical usage,'' \emph{TechRxiv}, Sep. 2023.

\bibitem{koh2023grounding}
J.~Y. Koh \emph{et~al.}, ``Grounding language models to images for multimodal generation,'' \emph{arXiv preprint arXiv:2301.13823}, Jun. 2023.

\bibitem{mildenhall2021nerf}
B.~Mildenhall, P.~P. Srinivasan, M.~Tancik, J.~T. Barron, R.~Ramamoorthi, and R.~Ng, ``{NeRF}: Representing scenes as neural radiance fields for view synthesis,'' \emph{Communications of the ACM}, vol.~65, no.~1, pp. 99--106, 2021.

\bibitem{takagi2023high}
Y.~Takagi and S.~Nishimoto, ``High-resolution image reconstruction with latent diffusion models from human brain activity,'' in \emph{Proceedings of the IEEE/CVF Conference on Computer Vision and Pattern Recognition}, 2023, pp. 14\,453--14\,463.

\bibitem{yang2020deep}
Y.~Yang \emph{et~al.}, ``Deep learning based antenna selection for channel extrapolation in {FDD} massive {MIMO},'' in \emph{International conference on wireless communications and signal processing (WCSP)}.\hskip 1em plus 0.5em minus 0.4em\relax IEEE, 2020, pp. 182--187.

\bibitem{wang2022transformer}
Y.~Wang \emph{et~al.}, ``Transformer-empowered {6G} intelligent networks: From massive {MIMO} processing to semantic communication,'' \emph{IEEE Wireless Commun.}, 2022.

\bibitem{gilbert2023semantic}
H.~Gilbert \emph{et~al.}, ``Semantic compression with large language models,'' 2023.

\bibitem{carta2023grounding}
T.~Carta \emph{et~al.}, ``Grounding large language models in interactive environments with online reinforcement learning,'' Sep. 2023.

\bibitem{du2023improving}
Y.~Du \emph{et~al.}, ``Improving factuality and reasoning in language models through multiagent debate,'' May 2023.

\end{thebibliography}

\section*{Biographies}

\textbf{Lina Bariah} (lina.bariah@ieee.org) is a Senior Researcher at Technology Innovation Institute, UAE. 

\textbf{Qiyang Zhao} (Qiyang.Zhao@tii.ae) is a Lead Researcher at Technology Innovation Institute, UAE. 

\textbf{Hang Zou} (Hang.Zou@tii.ae) is a Researcher at Technology Innovation Institute, UAE. 

\textbf{Yu Tian} (Yu.Tian@tii.ae) is a Researcher at Technology Innovation Institute, UAE. 

\textbf{Faouzi Bader} (Carlos-Faouzi.Bader@tii.ae) is the Director of Telecom Unit at Technology Innovation Institute. 

\textbf{M{\'e}rouane Debbah} (Merouane.Debbah@ku.ac.ae) is a Professor at Khalifa University, UAE. 

\end{document}